\documentclass{article}

\usepackage{microtype}
\usepackage{graphicx}
\usepackage{subfigure}
\usepackage{booktabs} 

\usepackage{amsmath}
\usepackage{etoolbox}

\usepackage{amssymb}
\usepackage{algorithmic}
\usepackage{forloop}
\usepackage{fancyhdr}
\usepackage{stmaryrd}
\usepackage{textcomp}
\usepackage{bm}

\usepackage{hyperref}

\usepackage[accepted]{icml2021}

\icmltitlerunning{Plug and Play, Model-Based Reinforcement Learning}

\begin{document}

\twocolumn[
\icmltitle{Plug and Play, Model-Based Reinforcement Learning}



\icmlsetsymbol{equal}{*}

\begin{icmlauthorlist}
\icmlauthor{Majid Abdolshah}{Deakin}
\icmlauthor{Hung Le}{Deakin}
\icmlauthor{Thommen Karimpanal George}{Deakin}
\icmlauthor{Sunil Gupta}{Deakin}
\icmlauthor{Santu Rana}{Deakin}
\icmlauthor{Svetha Venkatesh}{Deakin}
\end{icmlauthorlist}

\icmlaffiliation{Deakin}{Applied Artificial Intelligence Institute ($\mathrm{A}^2\mathrm{I}^2$), Deakin Uni-versity, Geelong, Australia}

\icmlcorrespondingauthor{Majid Abdolshah}{majid@deakin.edu.au}

\icmlkeywords{Machine Learning, ICML}

\vskip 0.3in
]



\printAffiliationsAndNotice{ } 

\begin{abstract}
Sample-efficient generalisation of reinforcement learning approaches have always been a challenge, especially, for complex scenes with
many components. In this work, we introduce Plug and Play Markov Decision
Processes, an object-based representation that allows zero-shot integration
of new objects from known object classes. This is achieved by representing
the global transition dynamics as a union of local transition functions,
each with respect to one active object in the scene. Transition dynamics
from an object class can be pre-learnt and thus would be ready to
use in a new environment. Each active object is also endowed with
its reward function. Since there is no central reward function, addition
or removal of objects can be handled efficiently by only updating
the reward functions of objects involved. A new transfer learning
mechanism is also proposed to adapt reward function in such cases.
Experiments show that our representation can achieve sample-efficiency
in a variety of set-ups.%
\end{abstract}

\section{Introduction}
The use of deep neural networks as function approximators has enabled
new functionalities for reinforcement learning to provide solutions
for more complex problems \citet{van2015blocks}. However, adaptively
scaling and generalising trained agents to larger and complex environments
remains challenging. Object Oriented Reinforcement Learning (OORL)
is an attempt to leverage scalability of the solutions by learning
to optimally behave with respect to some classes of objects in an
environment. This knowledge can later be generalised to other environments
in which the same classes of objects are present \citet{watters2019cobra}.

\begin{figure}[t]
\begin{center}
\includegraphics[width=0.7\columnwidth]{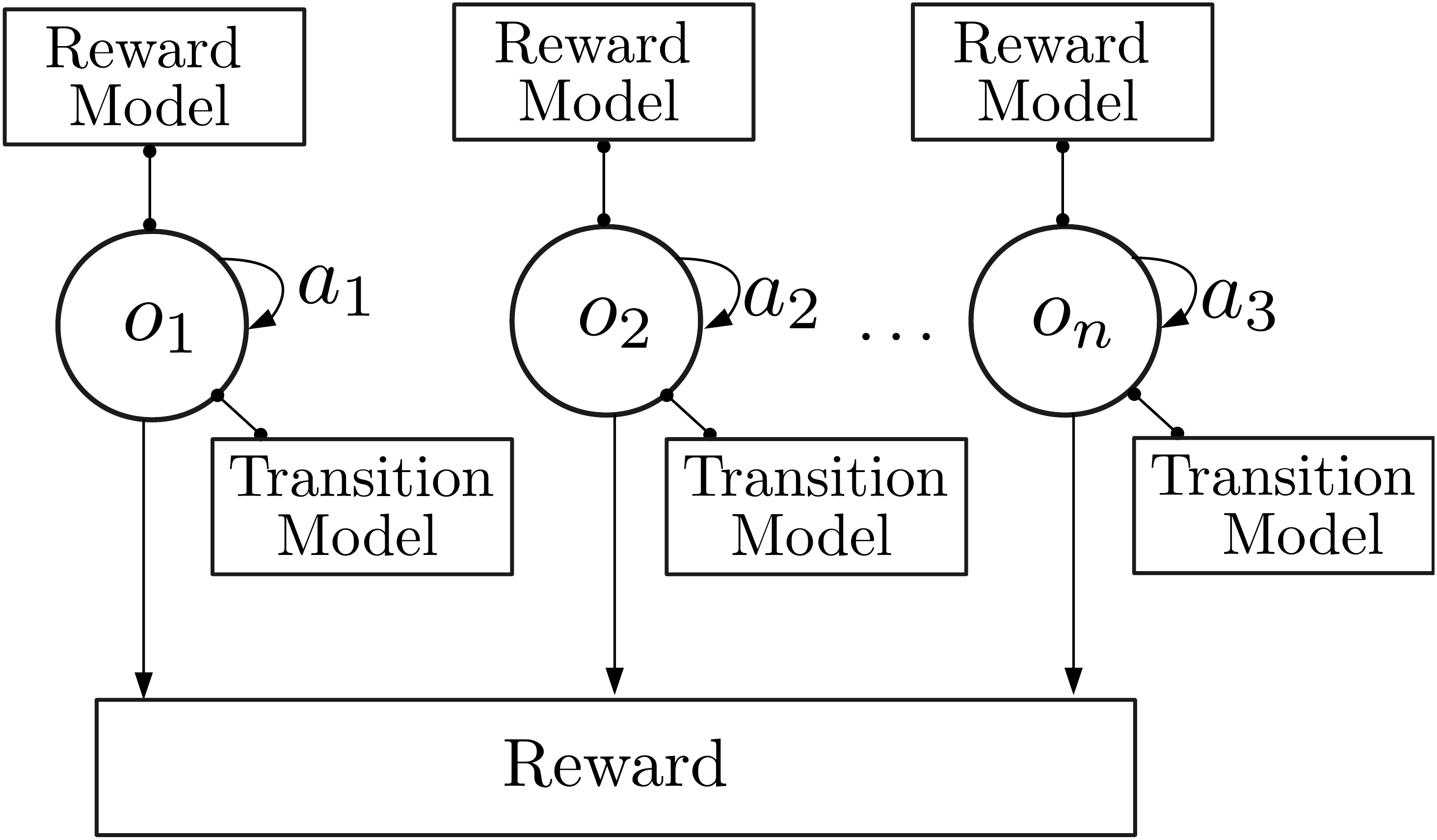} 
\end{center}
\caption{A Plug and play environment in which objects inherit a reward model
and a transition model form their class.}
 \label{fig:motiv}
\end{figure}

Most object-oriented reinforcement learning approaches aim to extract
objects in the environment by vision-based or vision-free approaches
\citet{watters2019cobra,kansky2017schema,keramati2018strategic}.
An agent learns how the environment ``works'' in a task-free exploration
phase by constructing a global transition model wherein the attributes
of objects from different classes can be manipulated (e.g. moving
objects). These global transition models are amalgamated representations
of objects (e.g. interaction networks, interaction graphs \citet{battaglia2016interaction}).
Global transition models enable a single agent to adjust object attributes
to obtain high rewards. However, they fail if objects are dynamically
introduced or removed from the scene. Major re-training of the global
transition model is required to make the model accurate with such change of scene. Other methods learn
the local transition models of the objects \citet{watters2019cobra,scholz2014physics} and pass the learnt properties of objects
to a single agent to perform an action to maximise its returns. However,
in such settings, objects are considered to be neutral and are not
allowed to perform any action or construct their own reward function.
As a result, such global reward models are also to be re-trained with
only slight change in the environment.
These difficulties motivate an alternative approach of \emph{``using
objects in the environment with a plug and play functionality''}.
We define three main requirements for a plug and play environment:
\begin{itemize}
\item Factorising the global transition model of the environment into local
transition models of the classes of objects that are present in the
environment,
\item Factorising object specific reward models instead of a single global
reward model; and 
\item Allowing adaptation of object specific reward models in the new environment.
\end{itemize}
Following the requirements of a plug and play approach, as the first
step, we eliminate the need for a single agent to adjust all objects.
Instead, independent objects inherit attributes from their class and
maintain their own local transition model and reward model. The global
transition dynamics is represented as a union of local transition
models, each with respect to one class of active objects in the scene.
Transition dynamics from an object class are pre-learnt and ready
for use in a new environment. Scenes can also be dynamically configured
with addition and removal of objects, and number of objects can be
arbitrarily large as they do not share a common reward model or a
transition model. Additionally, we develop a novel \textquoteleft trust
factor\textquoteright{} based transfer mechanism for reward functions
across environments. We term our approach as \emph{Plug and Play Reinforcement
Learning }(PaPRL). Figure \ref{fig:motiv} shows the overall structure
of a plug and play approach.

Experiments show that our representation achieves sample-efficiency
in a variety of set-ups from simple to complex environments. Building
upon the plug and play environments, objects can be arbitrarily added
or removed during the run time. To illustrate the effects of local
transition model of a class of objects, we consider two cases of (1)
learning the local transition models in an inexpensive and fast simulator
and then transfer/plug in a new environment (PaPRL-offline), and (2)
learning local transition model for each object during the run time
(PaPRL-online).

\section{Related Works}

This work builds upon three areas of related research:

\textbf{Physics-Based Reinforcement Learning}: In physics based reinforcement
learning, Deep Lagrangian Networks \citet{lutter2019deep} and Interaction
Networks \citet{battaglia2016interaction} aim to reason about the
way objects in complex systems interact. Deep Lagrangian Networks
(DeLaN) \citet{lutter2019deep} encode the physics prior in the form
of a differential in the network topology to achieve lower sample complexity and better extrapolation. However,
these works use a global transition model. Whilst they can naturally
handle arbitrary number of objects, they are not flexible in including
new classes of objects. Consequently, new class of objects require
complete retraining of the global transition model, and hence, this
may not be feasible in a plug and play setting in which the new classes
of objects may be introduced at any time.

\textbf{Object-Oriented Reinforcement Learning}: Object-oriented MDPs
\citet{diuk2008object} are an extension to the standard MDPs to construct
the global transition dynamics of the environment and develop an object-oriented
non-transferable reinforcement learning.  Following this approach some studies address the need for object-centric
understanding of the environment \citet{cobo2013object,garnelo2016towards,keramati2018strategic},
however, they are not naturally transferable to a new scene as the
objects' properties cannot be reused in a new environment. Keramati
et al. \citet{keramati2018strategic} aims to boost the performance
of agents in a large state space with sparse reward. Schema networks
\citet{kansky2017schema} use a vision system that detects and tracks
entities in an image. Given the entities, a self-transition variable
is defined to represent the probability that a position attribute
remains active in the next time step. While this self-transition variable
can be sufficient in some video games, it may not be suitable for
more complex physical environments with many attributes for objects.
Scholz et al. \citet{scholz2014physics} propose a A Physics-Based
model prior for object-oriented Reinforcement Learning (PBLR) with
a focus on Robotics applications. This representation defined a state-space
dynamics function in terms of the agent's beliefs over objects' inertial
parameters and the existence and parametrisation of physical constraints.
PBLR's approach is quite different and the solutions do not translate
into the settings of a Plug and Play framework in which every object
holds its own inherited properties. Model-based reinforcement learning
approaches such as COBRA \citet{watters2019cobra} aim to learn representations
of the world in terms of objects and their interactions and pass this
knowledge as a transition model to the single agent and with a global
reward model. Hence, this approach cannot be used in a plug and play
environment in which the objects are acting based on their own local
transition model and their own reward model.

\textbf{Multi-Agent Reinforcement Learning}: These approaches may
also be considered as related studies as in our proposed method, objects
can perform actions in the environment. However, the focus of multi-agent
methods is generally on improving the cooperation setting among agents
with same class of attributes to maximise a common long-term return,
and thus are not suitable for a plug and play environment where objects
can come from a variety of classes. MOO-MDP \citet{da2017moo} is
a study that combines the concept of object-oriented representation
and multi-agent reinforcement learning. This method is a model-free
algorithm but uses a global transition model to solve deterministic
multi-agent, object-oriented problems with a discrete action. The
extracted knowledge from objects are not transferable to a new environment.
Model-based multi-Agent reinforcement learning approaches such as
\citet{bargiacchi2020model}, aim to learn a model of the environment
and update a single reward model in a sample-efficient manner with
cooperation among agents. We however, are interested in modelling local transition dynamics, and we consider the agents to act independently. 

\textbf{Relations to Plug and Play Environments}: Based on the three
main requirements of a plug and play environment, physics-based reinforcement
learning approaches fail to fit into this setting as they only develop
a global transition model of the environment. Whilst some of the object-oriented
reinforcement learning approaches construct local transition models,
but they develop a single reward model that is incompatible with the
plug and play approach. Additionally, local transition models of objects
in such methods are not reusable in new environments as they are environment-specific.
Model-based multi-agent reinforcement learning methods generally work
with similar agents with same classes of attributes and the focus
of such methods is on improving the co-operation of agents with the
help of global transition model to achieve the highest return. Consequently,
such approaches cannot be applied into a plug and play environment.

\section{Plug and Play Markov Decision Processes}
Following the sequential decision-making problems, our plug and play
reinforcement learning framework is constructed based on MDPs and
object oriented MDPs. Plug and Play Markov Decision Process (PaP-MDP)
is described by $<\mathcal{\mathrm{{\mathcal{O}},\ }{C}},\mathcal{Z},\mathcal{{A}},\mathcal{{P}},\mathcal{{R}}>$,
where:
\begin{itemize}
\item $\mathcal{{O}}=\{o_{1},o_{2},\ldots,o_{O}\}$ is the set of objects
that are existing in the environment, 
\item $\mathcal{{C}=}\{c_{1},c_{2},\ldots,c_{C}\}$ is the set of classes
of objects that are present in the environment,
\item $\mathcal{Z}=\{z_{1},z_{2},\ldots,z_{O}\}$ is the set of booleans
that determines if an object is allowed to take an action in the environment:
$z_{i}=\{0,1\},\mathrm{and}\sum_{i\in\{1,\ldots,O\}}z_{i}=O^{\prime};1\leq O^{\prime}\leq O$,
\item $\mathcal{{A}}=\{a_{1},a_{2},\ldots,a_{O}\}$ is the set of valid
actions the objects in the environment $o_{i}\in\mathcal{O}$ may
take at every time-step if $z_{i}=1$, 
\item $\mathcal{{P}}=\{p_{1},p{}_{2},\ldots,p_{O}\}$ is the set of local
transition models for the objects in the environment $o_{i}\in\mathcal{O}$
if $z_{i}=1$; and
\item $\mathcal{{R}}$ is the global reward function. 
\end{itemize}
Different classes of objects have distinct attributes: $\mathrm{{att}}(c_{i})=\{e_{1}^{i},\ldots,e_{E_{i}}^{i}\},\forall i\in{1,\ldots C}$
and $E_{i}$ is the number of attributes for class $c_{i}$. These
attributes are inherited by new objects generated from class $c_{i}$.
To efficiently determine the class of each object, we define $\mathrm{C}(o_{i})=c_{k\in\{1\ldots C\}}$
that returns the class of an object. Attributes of objects may be
changed only by that specific object. If $z_{i}=1$, then $o_{i}$ can perform an action, we term these
objects as \textbf{\emph{active objects}}\emph{. }On the other hand,
if $z_{i}=0$, then $o_{i}$ is not permitted to perform any action
in the environment and we term these objects as \textbf{\emph{neutral
objects}} $\mathcal{N}=\cup o_{i},(\forall i\in\{1,\dots,O\}\land z_{i}=0)$
and $a_{i}=\emptyset$, a null action. Active objects can observe
the neutral objects' attributes. However, active objects are unaware
of each others' attributes and they do not interact directly with
each other. 

Given the definition of objects and attributes, the state of an object
$o_{i}$ in the environment is defined as:

\begin{equation}
o_{i}.s=[o_{i}.e_{1}^{\mathrm{C}(o_{i})},\ldots,o_{i}.e_{E}^{\mathrm{C}(o_{i})}]\label{eq:o_state}
\end{equation}
where the the ``dot'' notation denotes the attribute of an object.
The state of PaP-MDP is defined as the union of all states for different
objects that are present in the environment:

\begin{equation}
\mathcal{S}=\cup_{\forall o\in\mathcal{{O}}}\quad o.s\label{eq:o_state_2}
\end{equation}

\subsection{Local Transition Models \label{subsec:Local-Transition-Models}}

\begin{figure}[t]
\centering \includegraphics[width=0.8\columnwidth]{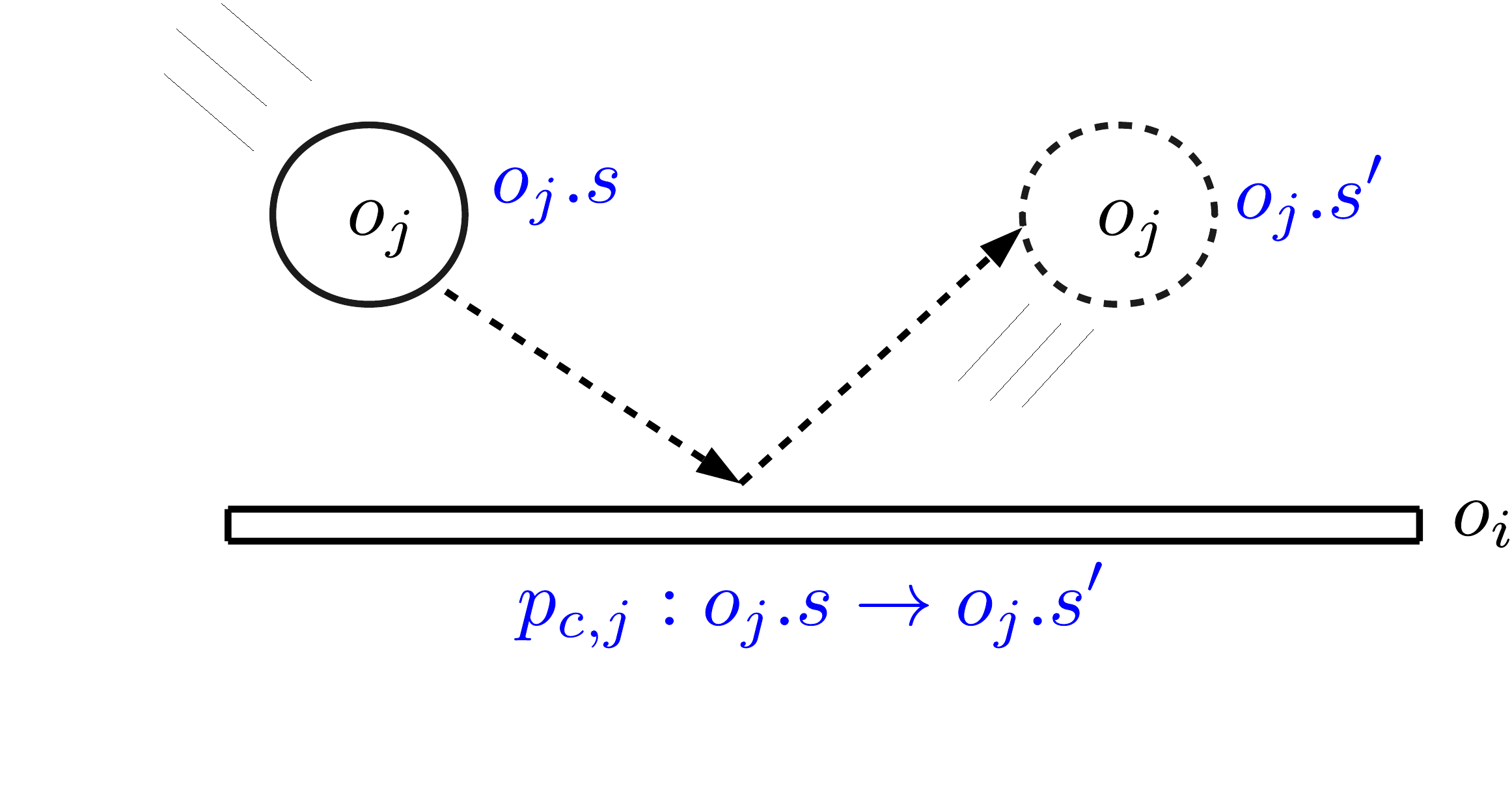}\caption{The interaction between $o_{i}$ and neutral object $o_{j}$ is modeled
by $p_{c,j}$.}
\label{fig:dynamic}
\end{figure}

Local transition models are defined for each class of active objects
in the environment. These local models act as dynamics models that
learn and encode the behaviour of different classes of active objects
as they interact with neutral objects. $p_{c,j}:o_{j}.s\rightarrow o_{j}.s^{\prime},\forall o_{j}\in\mathcal{N}$,
is an approximation of a local transition function for active object
$o_{i},c=\mathrm{C}(o_{i})$, that learns how to predict the next
state of the neutral object $o_{j}$ in the environment after they
interact. Active objects can observe the attributes of neutral objects,
hence, they can simply record the attributes of the neutral object
before and after the interaction, e.g. a collision can be determined
based on the distance of the objects. As an example, Figure \ref{fig:dynamic}
shows an active object $o_{i}$ before and after collision with a
neutral object $o_{j}$. $p_{c,j},c=\mathrm{C}(o_{i})$, the local
transition model, predicts $o_{j}.s^{\prime}$ given $o_{j}.s$ -
that is the state of the neutral object before the collision with
$o_{i}$. Linear function approximation may suffice to describe simple
interactions, for more complex and non-linear behaviours such as collision,
a non-linear function approximator is required. In this case, we use
a neural network $\mathcal{D}$ with weights $\theta$ trained with
a backpropagation algorithm to learn the interaction in a supervised
manner. The network is trained to minimise the MSE loss, using the
post-interaction state of the neutral object $o_{j}$ as label:

\[
\mathcal{L}_{j}(\theta)=\mathrm{\mathbb{E}}[(o_{j}.s^{\prime}-\mathcal{D}(o_{j}.s,\theta))^{2}]
\]

The active objects thus predict the post-interaction state of the
neutral object before interaction.

\subsection{Reward Learning Algorithm \label{subsec:Reward}}

Following PHYRE study \citet{bakhtin2019phyre}, we design our main
experimental environment to be \emph{``wait and see''}. Each active
object ($\forall o_{i}\notin\mathcal{N}$) is allowed to perform an
action only if interacting with a neutral object $o_{j}$. If $o_{i}$
and $o_{j}$ interact more than one time during an episode, $o_{i}$
is not allowed to perform more actions and will be treated as a neutral
object until it receives the  reward and completes the episode. 

\subsubsection{Reward Model \label{subsec:Reward-Model}}

\begin{figure}[t]
\centering{}\includegraphics[width=0.9\columnwidth]{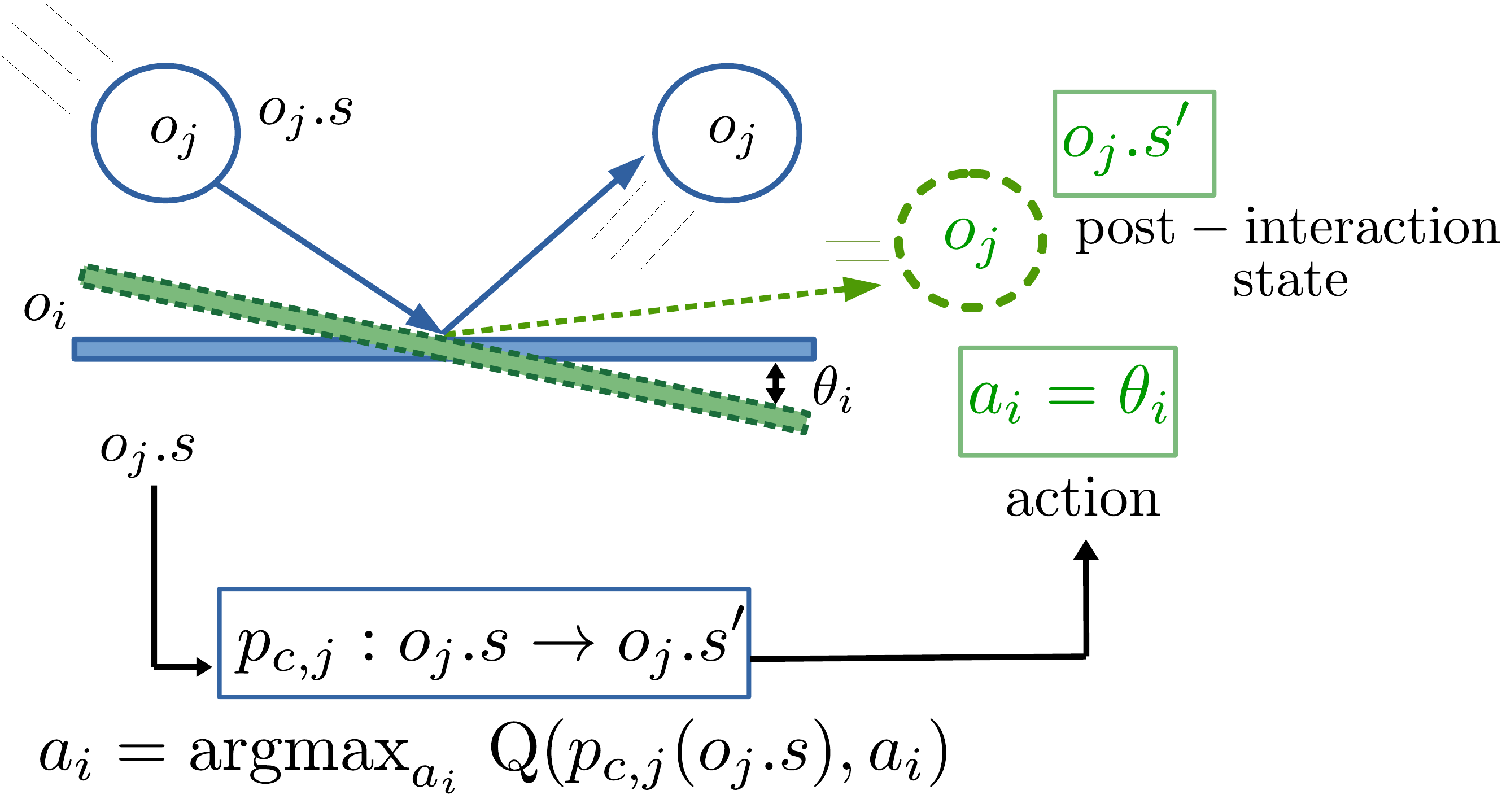}\caption{An example of interaction between active object $o_{i}$ and neutral
object $o_{j}$. As $o_{i}$ observes the attributes of $o_{j}$,
it performs a rotation action $a_{i}$ that results in the post-interaction
state $o_{j}.s^{\prime}$ that is the most favourable to return the
highest reward based on the trained rewarding model. Solid blue arrows
show the trajectory of $o_{j}$ if no action is taken by $o_{i}$.
Green dashed arrows shows the post-interaction trajectory of $o_{j}$
if action $a_{i}$ is performed.}
\label{fig:dynamic_2}
\end{figure}

We use a specific type of DQN to model the reward function in the
environment. Note that in the proposed problem, objects cannot perform
more than one action in a single episode and the time-steps are eliminated.
As a result, the Q-network of an active object represents a model
of its reward function. All active objects have one common goal, yet
they are not aware of each other's attributes. Given that, each object
collects a set of state-action-reward triplets after interacting with
a neutral object. However, given the trained local transition model
for each active object, we use triplets of the form \emph{``<post-interaction
state, action, reward>'}'. We argue that the learned local transition
model is a critical prior knowledge about the physics of an object
that can improve sample-efficiency of this approach. In a practical
sense, an active object can adjust itself by performing an action
that results in a post-interaction state which is more favourable
to achieve the highest reward. Consider the example of a ball as a~neutral
object ($o_{j}\in\mathcal{N}$) and an obstacle ($o_{i}\notin\mathcal{N}$)
as an  active object. Given the collision as a type of interaction
we are interested in, Figure \ref{fig:dynamic_2} shows that given
$p_{c,j},c=\mathrm{C}(o_{i})$, the local transition model, $o_{i}$
performs an action which results in desirable post-interaction state
that is expected to return the highest global reward:

\begin{equation}
a_{i}=\mathrm{argmax}_{a}o_{i}\mathrm{.Q}(p_{c,j}(o_{j}.s),a_{i}),\quad c=\mathrm{C}(o_{i})\label{eq:action}
\end{equation}
where $\mathrm{Q}$ is a neural network that is minimised by MSE loss
between the prediction and the observed reward:

\begin{equation}
\mathcal{L^{\prime}}_{j}(\theta)=\mathrm{E}[(r(\mathcal{S},\mathcal{A})-o_{i}.\mathrm{Q}(p_{c,j}(o_{j}.s),a_{i}))^{2}],\quad c=\mathrm{C}(o_{i}).\label{eq:rew_loss}
\end{equation}
Note that $r(\mathcal{S},\mathcal{A})$ is the global reward function
based on the environment's set of states $\mathcal{S}$ and object's
set of actions $\mathcal{A}$.

\subsubsection{Trust Factor}

A key part of our plug and play reinforcement learning framework is
derived from the concept of object oriented environments. As demonstrated
in reward learning section, each active object $o_{i}\notin\mathcal{N}$
constructs its own reward model $o_{i}\mathrm{.Q}$. However, $o_{i}$
also inherits a reward function from class $\mathrm{C}(o_{i})$ upon
initilisation, as the reward model is also one of the attributes of
that class. We term an inherited reward model for $o_{i}$ from class
$c=\mathrm{C}(o_{i})$, as $c.\mathrm{Q}.$ The inherited reward model
may not be the best choice as a new active object can be initialised
with different attributes such as position, length, weight, etc. We
impose no restrictions on $c.\mathrm{Q}$ and it can be a randomly
initialised neural network or any reward model from one of the previously
trained objects from that class. If the class reward model is selected
from one of the previously trained objects in the environment, a newly
initialised object may find the class reward function as a reasonable
starting model to partially rely on. To leverage the mixed use of
class and object models, we define a trust factor that intuitively
measures how accurately a reward model for an active object works.
The trust factor of approximated reward function $\mathrm{Q}$ at
iteration $t$ as:

\begin{equation}
\mathrm{TF}_{t}(\mathrm{Q})=(1+\sum_{k=1}^{K}|r(\mathcal{S},\mathcal{A})-\mathrm{Q}(p_{c,j}(o_{j}.s),a_{i})|)^{-1},\enskip c=\mathrm{C}(o_{i})\label{eq:trust}
\end{equation}
where $K$ is the number of episodes an object waits to calculate
its trust to the current reward model $\mathrm{Q}\in\{o_{i}.\mathrm{Q},\mathrm{C}(o_{i}).\mathrm{Q}\}$
that is currently being used. If the difference of the predicted and
the actual received rewards is considerable, the trust factor will
be low and the object attempts to use the alternative reward function
for the next round. Algorithm \ref{tab:tf} shows the trust factor
calculation.

\begin{algorithm}[t]
\caption{Trust Factor Calculation for Active Objects}

\begin{algorithmic}[1]

\STATE \textbf{Input:}

\STATE Active objects present in the environment: $\forall o_{i}\notin\mathcal{N}$,

\STATE Neutral objects present in the environment: $\forall o_{j}\in\mathcal{N}$, 

\STATE Reward models for all active objects $\mathrm{Q}\in\{o_{i}.\mathrm{Q},\mathrm{C}(o_{i}).\mathrm{Q}\},o_{i}\notin\mathcal{N}$, 

\STATE Initialize $K$, the step size and the trust threshold $h$
(see \ref{eq:trust}).

\STATE \textbf{Output: }Trust factors for all active objects at episode
$t$: $\mathrm{TF}_{t}(\mathrm{Q})$.

\FOR {$\forall o_i$}

\STATE $\mathrm{TF}_{t}(\mathrm{Q})=(1+\sum_{k=1}^{K}|r(\mathcal{S},\mathcal{A})-\mathrm{\mathrm{Q}}(p_{c,j}(o_{j}.s),a_{i})|)^{-1}$

\IF {$\mathrm{TF}_t(\mathrm{Q}) \leq h$}

\STATE $\mathrm{\mathrm{Q}=Replace}(o_{i}.\mathrm{Q},\mathrm{C}(o_{i}).\mathrm{Q})\quad$\emph{//Replace
the current reward model with the other available reward model}

\ENDIF

\ENDFOR

\end{algorithmic}

\label{tab:tf}
\end{algorithm}

\begin{figure}[h]
\begin{centering}
\includegraphics[scale=0.30]{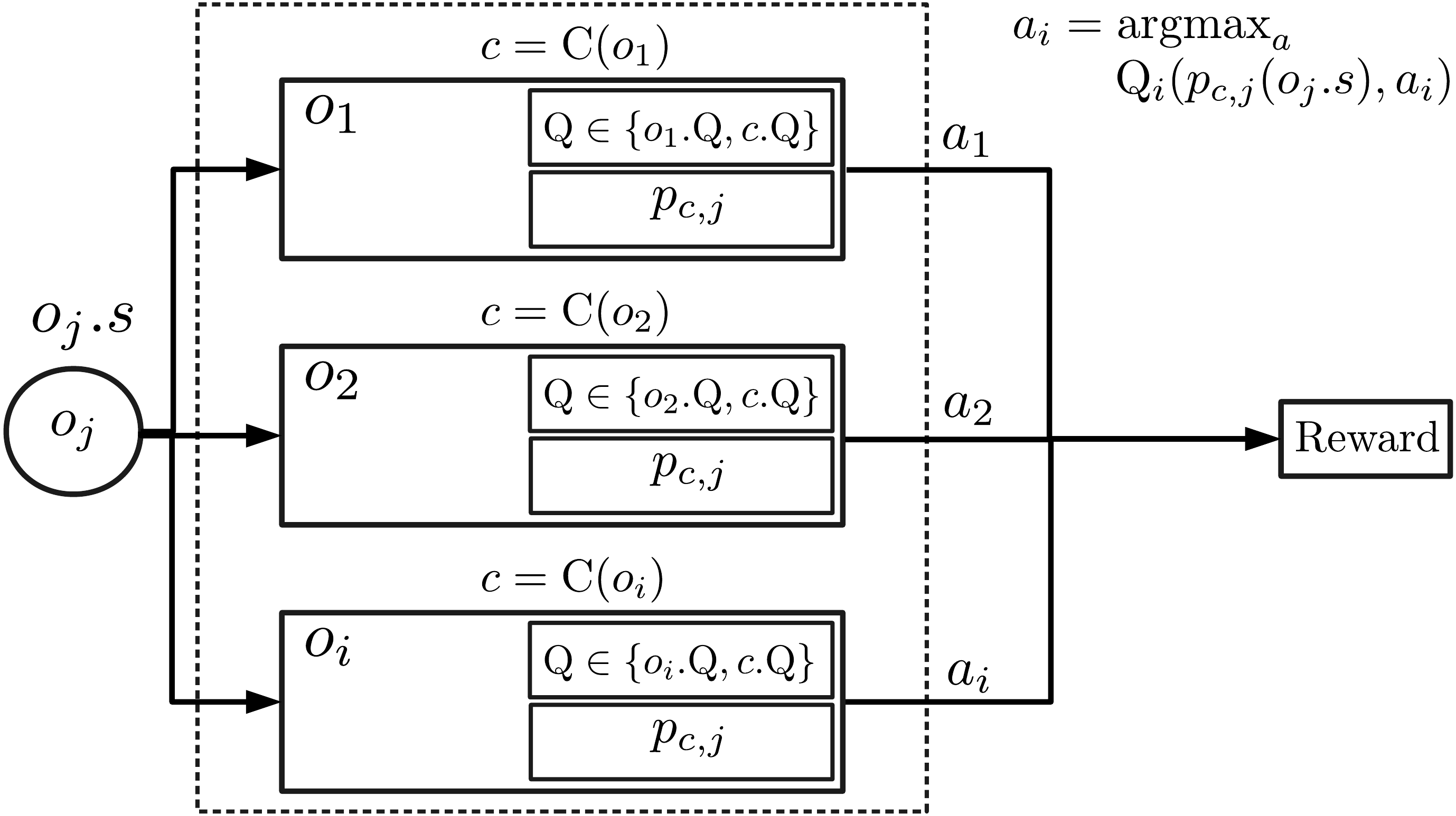}
\end{centering}
\caption{Plug and Play Reinforcement Learning}
\label{fig:framework}
\end{figure}
Figure \ref{fig:framework} shows the overview of our proposed plug
and play reinforcement learning framework in which all the objects
inherit their class attributes and attempt to maximise their own constructed
reward. Algorithm \ref{tab:method} outlines the pseudo code of plug
and play reinforcement learning.

\begin{algorithm}[t]
\caption{Plug and Play Reinforcement Learning Algorithm}

\begin{algorithmic}[1]

\STATE \textbf{Input:}

\STATE Active objects present in the environment: $\forall o_{i}\notin\mathcal{N}$,

\STATE Neutral objects present in the environment: $\forall o_{j}\in\mathcal{N}$,

\STATE Reward models for all active objects $\mathrm{Q}\in\{o_{i}.\mathrm{Q},\mathrm{C}(o_{i}).\mathrm{Q}\},\forall o_{i}\notin\mathcal{N}$,

\STATE Initialize $K$, the step size and the trust threshold $h$
(see \ref{eq:trust}).

\STATE \textbf{Output: }Updated estimates of $\mathrm{Q}\in\{o_{i}.\mathrm{Q},\mathrm{C}(o_{i}).\mathrm{Q}\},\forall o_{i}\notin\mathcal{N}$.

\FOR {$\forall o_j,\ o_j \in \mathcal{N}$}

\FOR {$\forall o_i,\ o_i \notin \mathcal{N}$}

\IF {$\mathrm{interact}(o_j,o_i)$}

\STATE $a_{i}=\mathrm{argmax}_{a_{i}}\mathrm{Q}(p_{c,j}(o_{j}.s),a_{i}),\quad c=\mathrm{C}(o_{i})$.

\STATE Construct the $<o_{j}.s^{\prime}$, $a_{i}$, $r(\mathcal{S},\mathcal{A})>$
and append to the observation set.

\STATE Sample a random minibatch of triplets and update $\mathrm{Q}$
with minimising Loss function \ref{eq:rew_loss}.

\IF {$\mathrm{TF}_t(\mathrm{Q}) \leq h$}

\STATE $\mathrm{\mathrm{Q}=Replace}(o_{i}.\mathrm{Q},\mathrm{C}(o_{i}).\mathrm{Q})\quad$
\emph{//Replace the current reward (Algorithm 1)}

\ENDIF

\ENDIF

\ENDFOR

\ENDFOR

\end{algorithmic}

\label{tab:method}
\end{algorithm}

\section{Experiments}
\begin{figure}
\begin{centering}
\begin{minipage}[t]{0.5\columnwidth}%
\noindent \begin{center}
\includegraphics[scale=0.23]{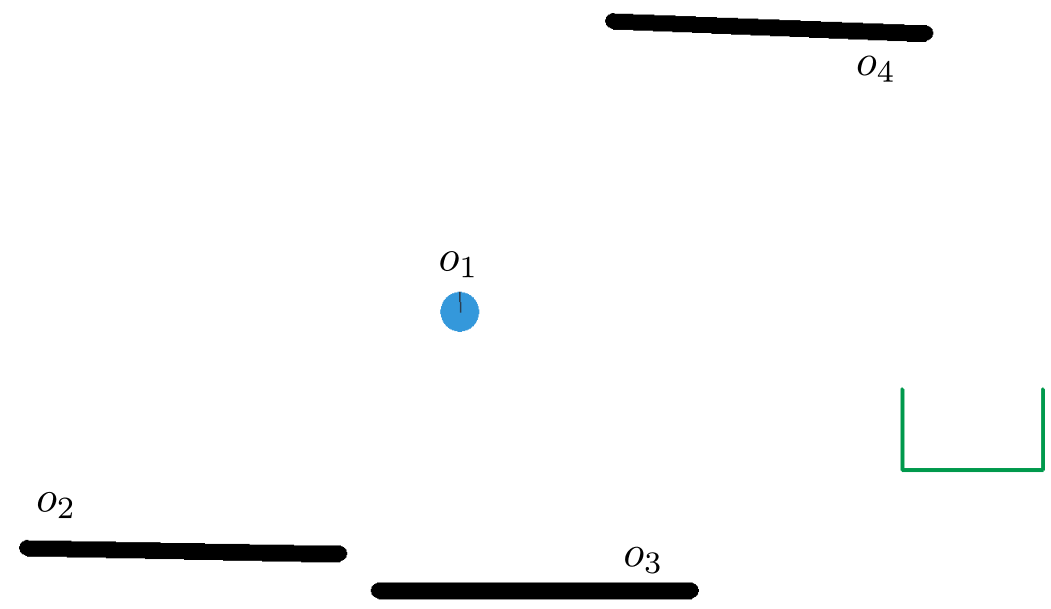}
\par\end{center}%
\end{minipage}
\par\end{centering}
\begin{raggedright}
(a) Rotating Wall active objects and the neutral-ball.
\par\end{raggedright}
\begin{centering}
\begin{minipage}[t]{0.5\columnwidth}%
\noindent \begin{center}
\includegraphics[scale=0.23]{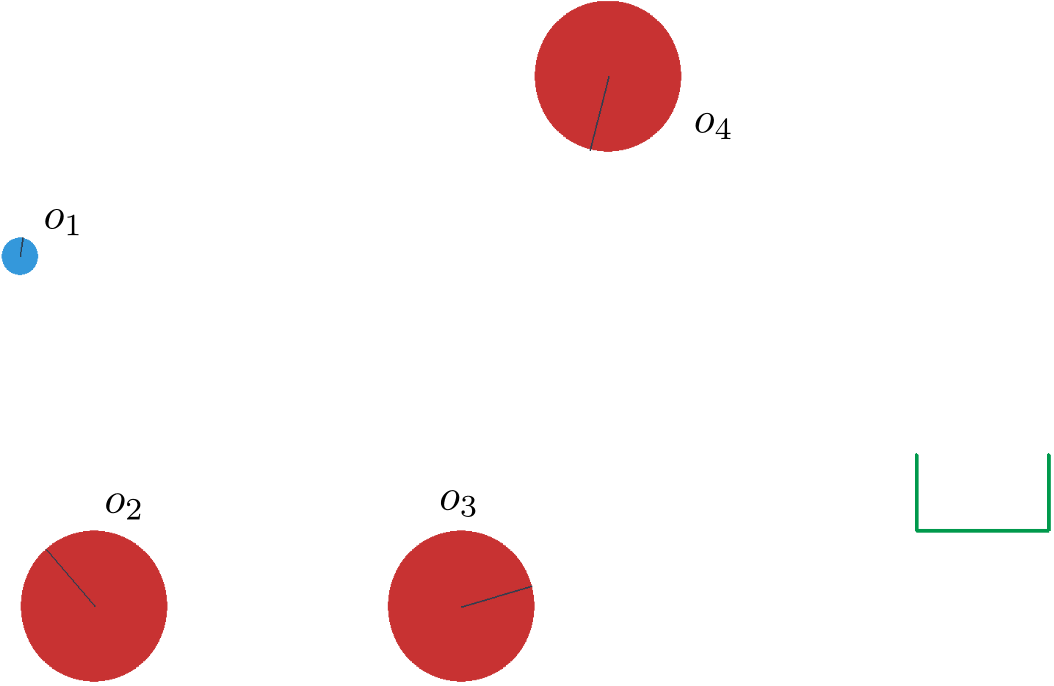}
\par\end{center}%
\end{minipage}
\par\end{centering}
\begin{raggedright}
(b) Arc Wall active objects (red) and the neutral-ball (blue).
\par\end{raggedright}
\begin{raggedright}
\caption{A sample experiment with active and neutral objects $\{o_{1}\}\in\mathcal{N},\{o_{2},o_{3},o_{4}\}\protect\notin\mathcal{N}$.
Active objects receive the highest reward if the neutral-ball falls
in the green basket.}
\par\end{raggedright}
\centering{}\label{fig:exp1}
\end{figure}

\begin{figure*}[t]
\begin{raggedright}
\begin{minipage}[t]{0.21\paperwidth}
\begin{center}
\includegraphics[width=0.8\textwidth]{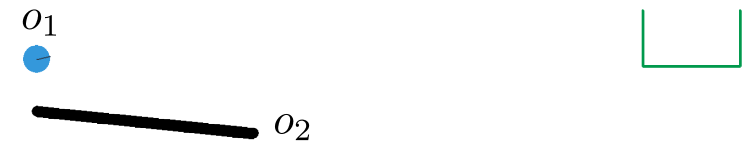}
\end{center}
\begin{center}
(a) An environment with $\{o_{1}\}\in\mathcal{N},\{o_{2}\}\notin\mathcal{N}$
as rotating wall.
\end{center}%
\end{minipage}%
\begin{minipage}[t]{0.21\paperwidth}%
\begin{center}
\includegraphics[width=0.2\paperwidth]{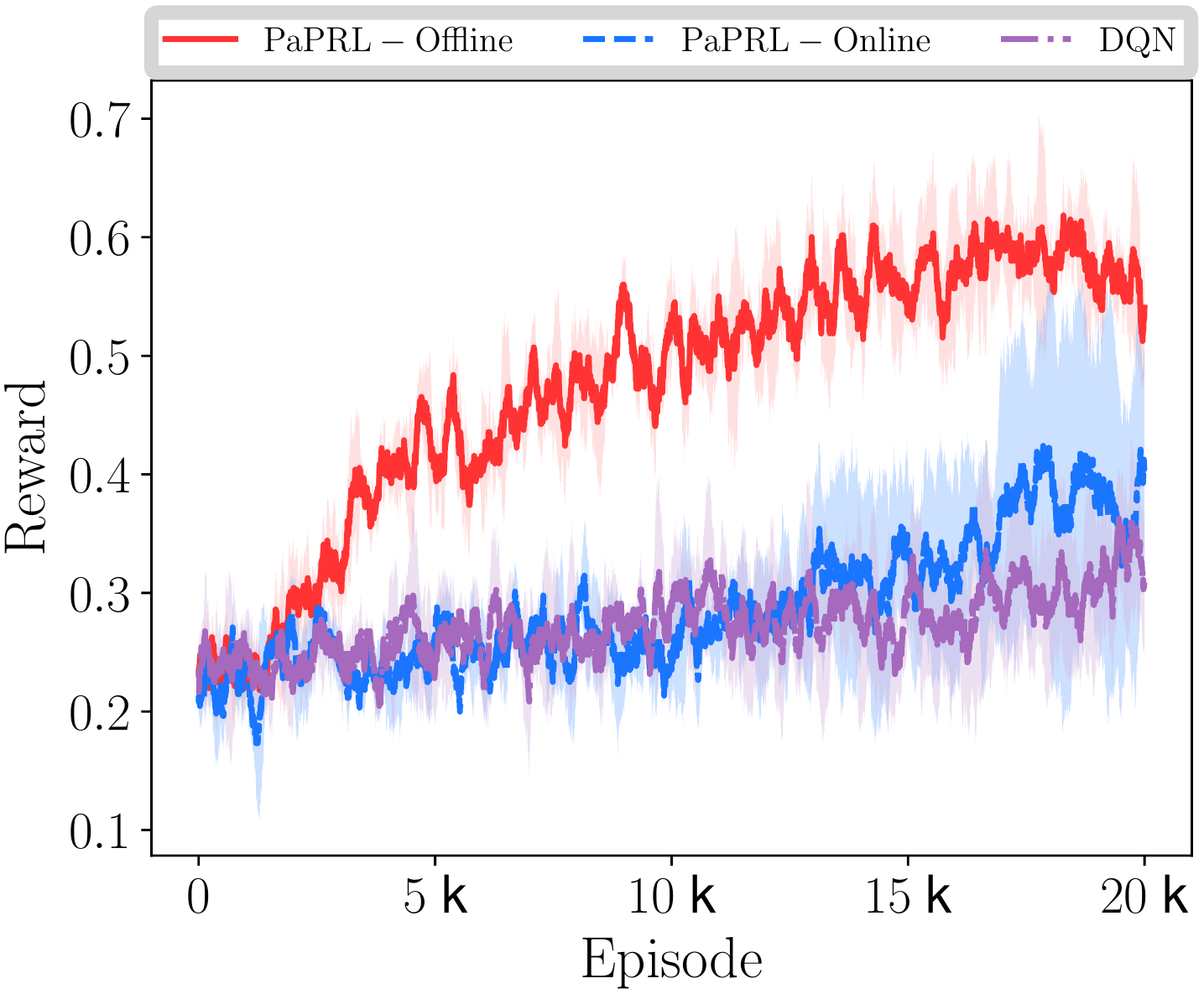}
\end{center}
\begin{center}
(b) Obtained reward from the environment.
\end{center}%
\end{minipage}%
\begin{minipage}[t]{0.21\paperwidth}%
\begin{center}
\includegraphics[width=0.2\paperwidth]{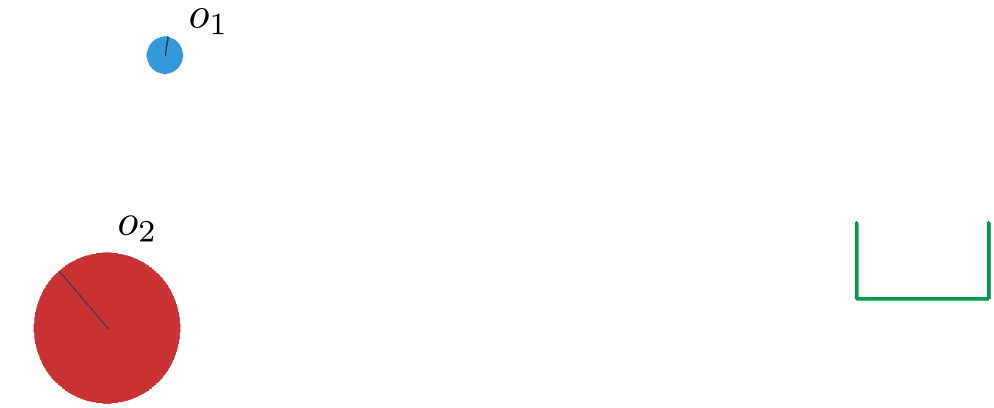}
\end{center}
\begin{center}
(c) An environment with $\{o_{1}\}\in\mathcal{N},\{o_{2}\}\notin\mathcal{N}$
as the arc wall.
\end{center}%
\end{minipage}%
\begin{minipage}[t]{0.21\paperwidth}%
\begin{center}
\includegraphics[width=0.2\paperwidth]{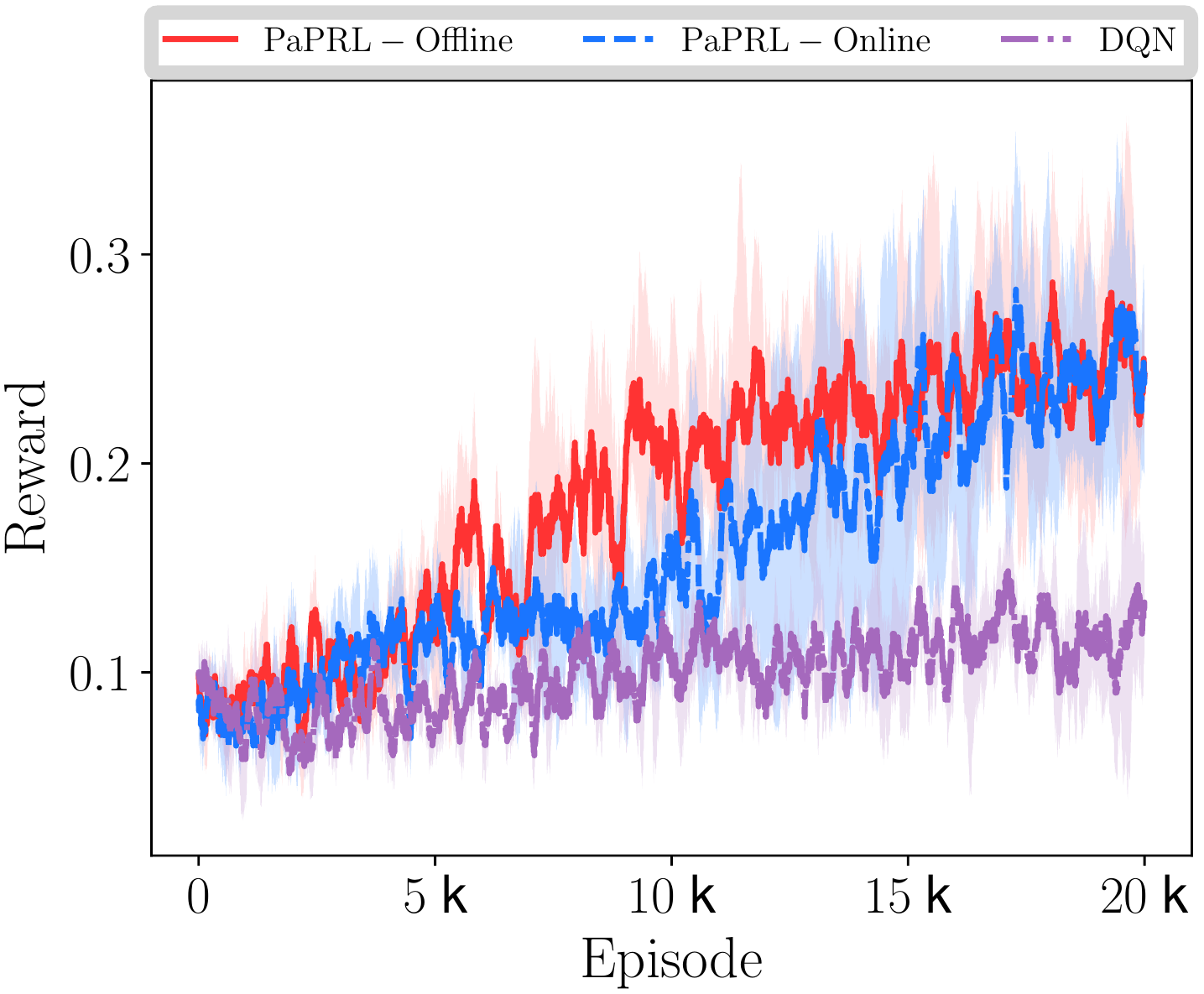}
\end{center}
\begin{center}
(d) Obtained reward from the environment.
\end{center}%
\end{minipage}
\end{raggedright}
\begin{raggedright}
\begin{minipage}[t]{0.2\paperwidth}%
\begin{center}
\includegraphics[width=0.8\textwidth]{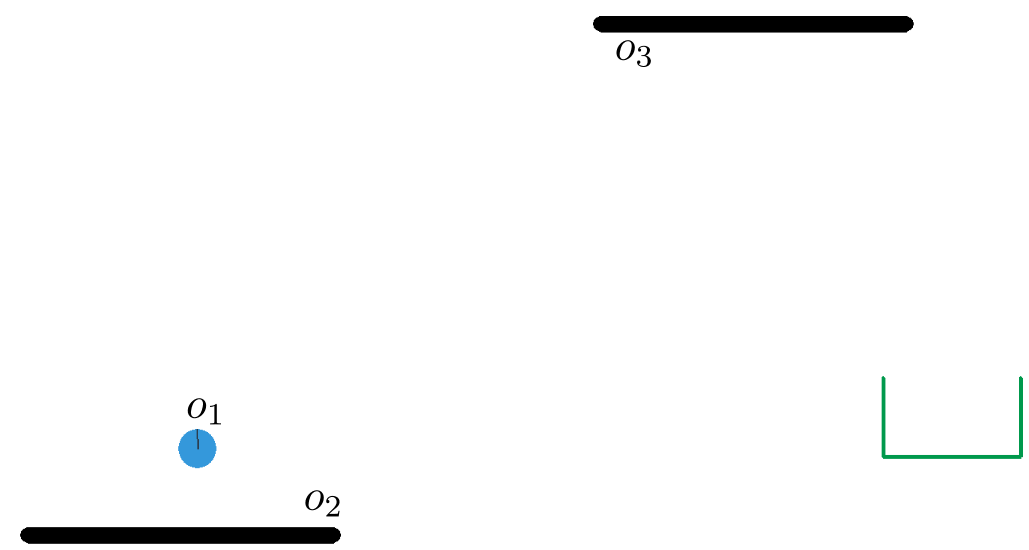}
\end{center}
\begin{center}
(e) An environment with $\{o_{1}\}\in\mathcal{N},\{o_{2},o_{3}\}\notin\mathcal{N}$
with two rotating walls.
\end{center}%
\end{minipage}%
\begin{minipage}[t]{0.23\paperwidth}%
\begin{center}
\includegraphics[width=0.9\textwidth]{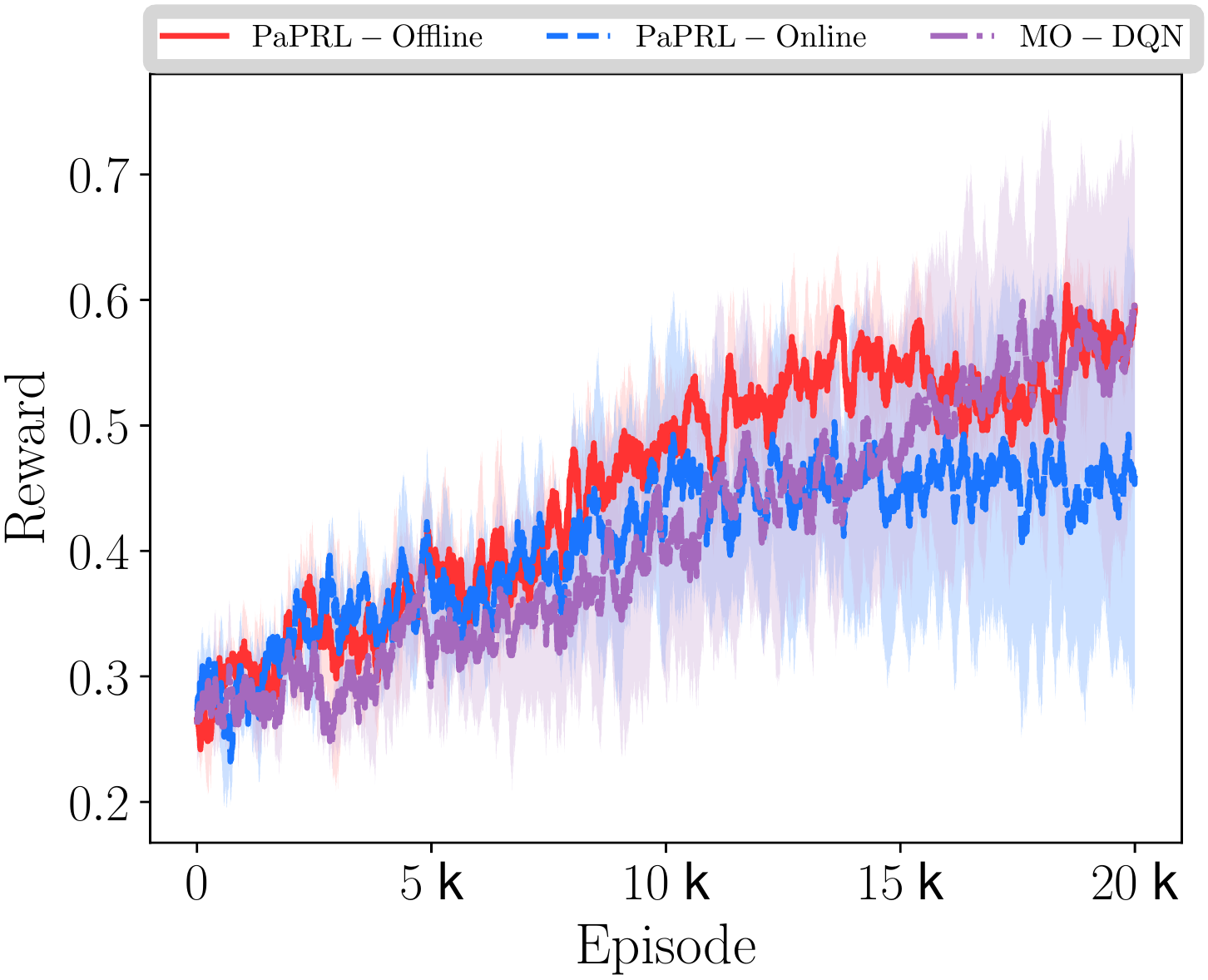}
\end{center}
\begin{center}
(f) Obtained reward from the environment.
\end{center}%
\end{minipage}%
\begin{minipage}[t]{0.2\paperwidth}%
\begin{center}
\includegraphics[width=0.8\textwidth]{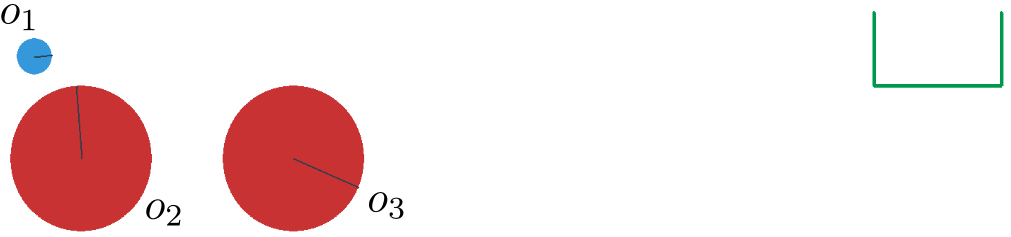}
\end{center}
\begin{center}
(g) An environment with $\{o_{1}\}\in\mathcal{N},\{o_{2},o_{3}\}\notin\mathcal{N}$
as arc walls.
\end{center}%
\end{minipage}%
\begin{minipage}[t]{0.23\paperwidth}%
\begin{center}
\includegraphics[width=0.9\textwidth]{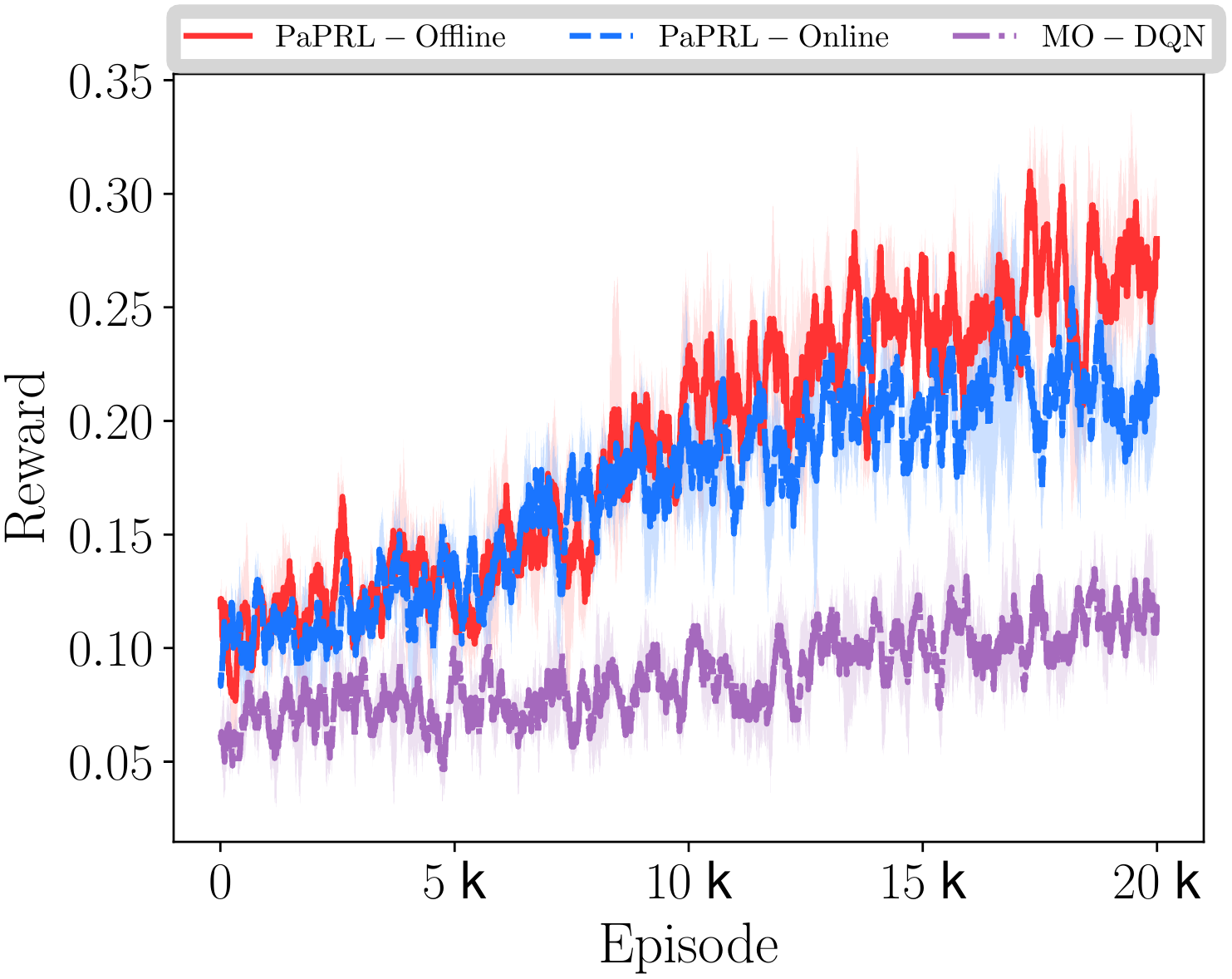}
\end{center}
\begin{center}
(h) Obtained reward from the environment.
\end{center}%
\end{minipage}
\end{raggedright}
\caption{Single (first row) and many (second row) active objects (rotating
wall and arc wall) in the environment with a comparison of PaP-RL
and DQN.}
\label{fig:exp2}
\end{figure*}

In this section, we demonstrate the performance of PaP-RL in a simple
plug and play reinforcement learning platform\footnote{Codes and videos are available in supplementary materials.},
following which, we present detailed empirical comparisons against
the baselines. 

\subsection{Experimental Setting\label{subsec:Settings-of-the}}

In our experiments, we introduce two types of objects: active objects
( Rotating Wall and Arc Wall) that can perform actions, and a neutral
object (Neutral-ball) that cannot. The two classes of active objects
have their own attributes that can be adjusted by performing the required
action to maximise the expected return.

\subsubsection{Rotating Wall}

The rotating wall class of active objects are constructed by the following
attributes: 

\[
\mathrm{att}(C_{w})=[f_{w},e_{w},\theta_{w}],
\]
where $f_{w}\in[0.4,0.9]$ is the friction coefficient of the wall
, $e_{w}\in[0.4,0.9]$ is the elasticity coefficient of the wall,
and $\theta_{w}\in[-\frac{\pi}{10},\frac{\pi}{10}]$ is the degree
of rotation for the wall. 

\subsubsection{Arc Wall}

The arc wall class of active objects are designed by the following
attributes: 

\[
\mathrm{att}(C_{a})=[f_{a},e_{a},\dot{v}_{a}],
\]
where $f_{a}\in[0.4,0.9]$ is the friction coefficient of the wall,
$e_{a}\in[0.4,0.9]$ is the elasticity coefficient of the wall, and
$\dot{v}_{a}\in[-50,50]$ is the angular velocity of the arc wall.

\subsubsection{Neutral-ball}

We assume the Neutral-ball class of objects have the following attributes: 

\[
\mathrm{att}(C_{b})=[v_{x},v_{y},\dot{v_{b}},\theta_{b},f_{b},e_{b},m_{b}],
\]
where $v_{x}$: velocity in x-axis, $v_{y}$: the velocity in y-axis,
$\dot{v_{b}}$: the angular velocity, $\theta_{b}$: the angle of
movement, $f_{b}\in[0.4,0.9]$: friction coefficient of the ball,
$e_{b}\in[0.4,0.9]$ elasticity coefficient of the ball, and $m_{b}\in[5,25]$:
mass of the ball. As discussed before, the post-interaction state
of the ball as a neutral object needs to be recorded in order to train
the local transition models of each active object. Hence, $o_{j}.s^{\prime}$
can be defined as:

\[
o_{j}.s^{\prime}=[v_{x}^{\prime},v_{y}^{\prime},\dot{v}_{b}^{\prime},\theta_{b}^{\prime}],
\]
where $v_{x}^{\prime}$: velocity in x-axis after interaction, $v_{y}^{\prime}$:
the velocity in y-axis after interaction, $\dot{v}_{b}^{\prime}$:
the angular velocity after interaction, $\theta_{b}^{\prime}$: the
angle of movement after interaction.

\subsubsection{Basket-Ball Platform }

We design a simple platform with one neutral object (blue ball) and
several active objects (either rotating or arc walls). The active
objects can perform actions to allow the blue ball to reach the basket.
A smooth reward function returns the maximum reward if the neutral-ball
moves into the desired location (e.g. a Basket). The active object
$o_{i}$ receives a reward after interaction with neutral object $o_{j}$
as follows:

\begin{equation}
r(o_{j}.s^{\prime},a_{i})=\begin{cases}
1, & \text{if}\ o_{j}\ \mathrm{in\ Basket}\\{}
[\mathcal{\mathrm{Min}}(d)]^{-1}, & \text{otherwise}
\end{cases}
\end{equation}
where $d$ is the distance of the ball with the center of the basket
at each time-frame, $o_{j}.s^{\prime}$ is the post-interaction state
of $o_{j}$, and $a_{i}$ is the action taken by $o_{i}$ to adjust
its attributes. Figure \ref{fig:exp1} illustrates an example of this
platform with two different classes of active objects. At every episode
of the environment, a neutral object is generated in a random position
and the active objects are required to adjust their attributes by
performing the proper action. Note that every active object is only
allowed to perform a single action in an episode and wait until it
receives the global reward from the environment. Further implementation details with more experiments are available
in supplementary materials.

\begin{figure*}[t]
\centering{}\includegraphics[width=0.6\paperwidth]{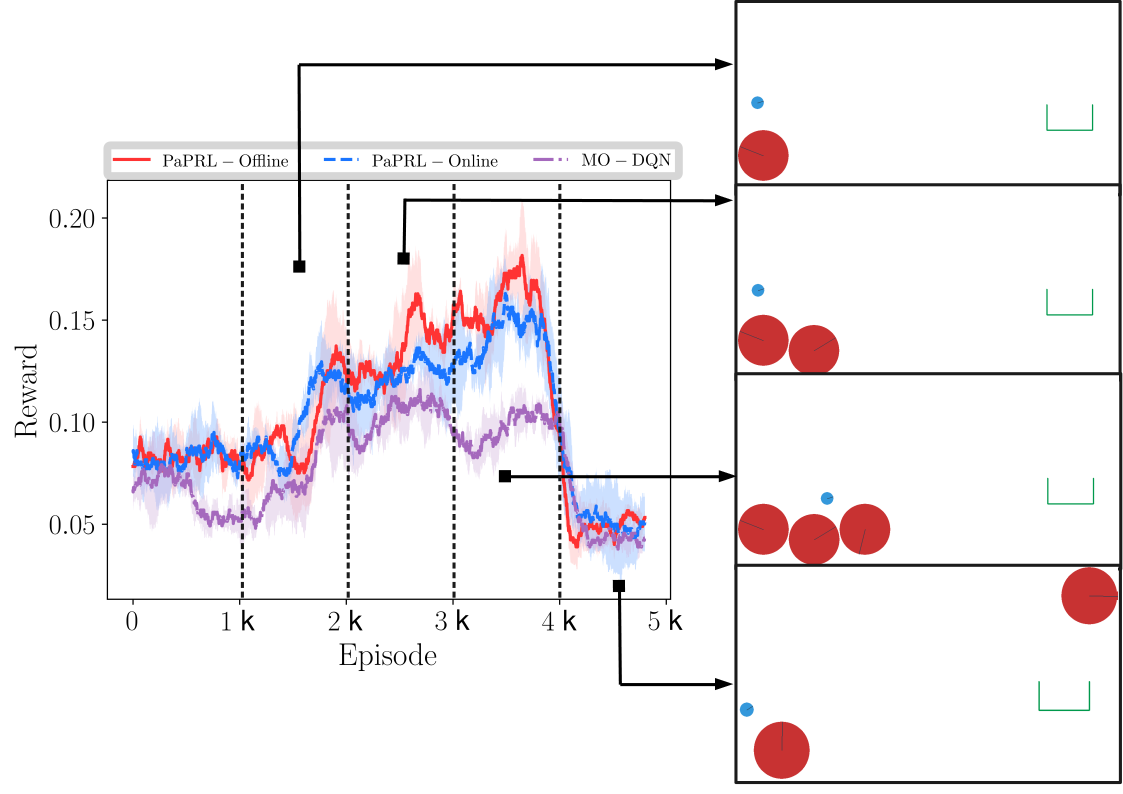}\caption{Effects of adding/removing active objects (red arc walls) on the reward
for PaPRL and MO-DQN.}
\label{fig:exp5}
\end{figure*}

\subsection{Single Active Object \label{subsec:Single-Active-Object}}

We start with an environment with a single active object and we proceed
to more complex environments. Our first experiment is based on a rotating
wall active object and a neutral-ball object. 

\subsubsection{Baselines}

To compare our proposed method with other related approaches, we show
the results of following approaches:
\begin{itemize}
\item \textbf{DQN as implemented in PHYRE} \citet{bakhtin2019phyre}: For
this baseline, we construct the Q-network with state-action pairs
and the returned reward as the target. The state in this case, is
the state of the neutral objects $o_{j}.s$ that is about to interact
with an active object $o_{i}$ that takes an action $a_{i}$ which
is expected to result in the highest return. Similar to PHYRE framework
that deals with continuous action space, we sample $10,000$ actions
at each episode and choose the one with the highest expected reward.
\item \textbf{PaP-Online:} PaP-RL with no pre-trained local transition function,
hence it learns the local transition models during the run time by
observing prior- and post-interaction states.
\item \textbf{PaP-Offline:} PaP-RL with pre-trained local transition function
from an inexpensive and fast simulator with no reward.
\end{itemize}
Figure \ref{fig:exp2} (first row) shows the results of single active
object experiments. Our offline and online PaP-RL methods outperforms
DQN, with the offline version being more sample-efficient as it relies
on a pre-trained local transition model. Whereas, the DQN approach
that only relies on prior-interaction of the neutral object and does
not benefit from local transition models have a slower rate of improvement.

\subsection{Many Active Objects}

We now extend the problem to incorporate many active objects $o_{i}\notin\mathcal{N}$
in the environment. 

\subsubsection{Baselines}

To compare our proposed method with other related approaches, we are
using two baselines as follows:
\begin{itemize}
\item \textbf{MO-DQN in PHYRE} \citet{bakhtin2019phyre}: We extended the
implemented DQN for every active object in the environment and we
call it Multi-Object DQN (MO-DQN). Following this approach, all the
active objects in the environment independently maintain a Q-network
based on state-action pairs and the returned reward as explained in
the settings of the experiments. Hence, at every episode, $o_{i}$
selects the action that maximises the expected reward based on its
constructed reward model.
\item \textbf{PaP-Online:} Similar to Single Active Object experiments,
we use PaP-RL with no pre-trained local transition function, hence
every active object in the environment learns the localy transition
models by observing prior- and post-interaction states, and its own
reward model independently.
\end{itemize}
Figure \ref{fig:exp2} (second row) shows the results of our experiment
with many active objects in the environment and it confirm that PaP-RL
outperforms DQN with higher numbers of active objects in the environment,
with the offline version being better as expected.

\subsection{Adding/Removing Objects}

As we have explained before, PaP-MDP is defined to incorporate addition/removal of active objects in the environment. Figure \ref{fig:exp5}
shows the effects of adding/removing objects in the environment during
the run time. We compare PaPRL with MO-DQN method by adding new objects
after 1000, 2000, and 3000 episodes. To illustrate the effects of
removal, two objects are removed at the 4000th episode. Figure \ref{fig:exp5}
shows that both PaPRL and DQN experience a higher rate of improvement
when a new object is added to the environment (with an exception of
DQN after episode 3000). The reason for this boost of improvement
is that the newly added active objects are likely to prevent the neutral
ball to leave the environment by violating the environment boundaries
and receiving a low reward. Hence, the neutral ball is given more
chances for (possible) interactions with new active objects to fall
in the basket. Figure \ref{fig:exp5} also shows that PaPRL-offline
outperforms other baselines as it uses the pre-trained local transition
models.

\section{Conclusion}
In this paper, we proposed the Plug and Play Markov Decision Processes
to introduce a plug and play, object-centric reinforcement learning
approach. In our proposed plug and play approach, independent objects
inherit attributes from their class and maintain their own local transition
model and reward model. Accordingly, the global transition dynamics
is represented as a union of local transition models, each with respect
to one class of active objects in the scene. We also showed that in
this framework, scenes can also be dynamically configured with addition
and removal of objects, and number of objects can be arbitrarily large.
Our experimental results prove sample-efficiency of our approach compared
to other related methods.

\bibliography{example_paper}
\bibliographystyle{icml2021}

\end{document}